\documentclass[letterpaper]{article} 
\usepackage{booktabs}
\usepackage{aaai25}  
\usepackage{times}  
\usepackage{helvet}  
\usepackage{courier}  
\usepackage[hyphens]{url}  
\usepackage{graphicx} 
\usepackage{natbib}  
\usepackage{caption} 
\usepackage{algorithm}
\usepackage{algorithmic}
\usepackage{amsmath}
\usepackage{amssymb}
\usepackage{amsmath}
\usepackage{booktabs}
\usepackage{multirow}
\usepackage[table,xcdraw]{xcolor}
\usepackage{newfloat}
\usepackage{listings}
\DeclareCaptionStyle{ruled}{labelfont=normalfont,labelsep=colon,strut=off} 
\floatstyle{ruled}
\newfloat{listing}{tb}{lst}{}
\floatname{listing}{Listing}
\title{Robust SAM: On the Adversarial Robustness of Vision Foundation Models}
\author{
    Jiahuan Long\textsuperscript{\rm 1, 2},
    Zhengqin Xu\textsuperscript{\rm 1},
    Tingsong Jiang\textsuperscript{\rm 2},
    Wen Yao\textsuperscript{\rm 2}\equalcontrib,
    Shuai Jia\textsuperscript{\rm 1}, \\
    Chao Ma\textsuperscript{\rm 1}\equalcontrib,
    Xiaoqian Chen\textsuperscript{\rm 2}
}
\affiliations{
    \textsuperscript{\rm 1}MoE Key Lab of Artificial Intelligence, AI Institute, Shanghai Jiao Tong University, \\
    \textsuperscript{\rm 2}Intelligent Game and Decision Laboratory, Defense Innovation Institute, Chinese Academy of Military Science, 
\\
    jiahuanlong@sjtu.edu.cn, wendy0782@126.com, chaoma@sjtu.edu.cn
}

\usepackage{bibentry}

\begin{document}

\maketitle

\begin{abstract}
The Segment Anything Model (SAM) is a widely used vision foundation model with diverse applications, including image segmentation, detection, and tracking. 
Given SAM's wide applications, understanding its robustness against adversarial attacks is crucial for real-world deployment. However, research on SAM's robustness is still in its early stages. 
Existing attacks often overlook the role of prompts in evaluating SAM's robustness, and there has been insufficient exploration of defense methods to balance the robustness and accuracy.
To address these gaps, this paper proposes an adversarial robustness framework designed to evaluate and enhance the robustness of SAM. 
Specifically, we introduce a cross-prompt attack method to enhance the attack transferability across different prompt types. 
Besides attacking, we propose a few-parameter adaptation strategy to defend SAM against various adversarial attacks. 
To balance robustness and accuracy, we use the singular value decomposition (SVD) to constrain the space of trainable parameters, where only singular values are adaptable. 
Experiments demonstrate that our cross-prompt attack method outperforms previous approaches in terms of attack success rate on both SAM and SAM 2.
By adapting only 512 parameters, we achieve at least a 15\% improvement in mean intersection over union (mIoU) against various adversarial attacks. 
Compared to previous defense methods, our approach enhances the robustness of SAM while maximally maintaining its original performance. 
\end{abstract}

\section{Introduction}

\begin{figure}[t]
\centering
\includegraphics[width=1.0\linewidth]{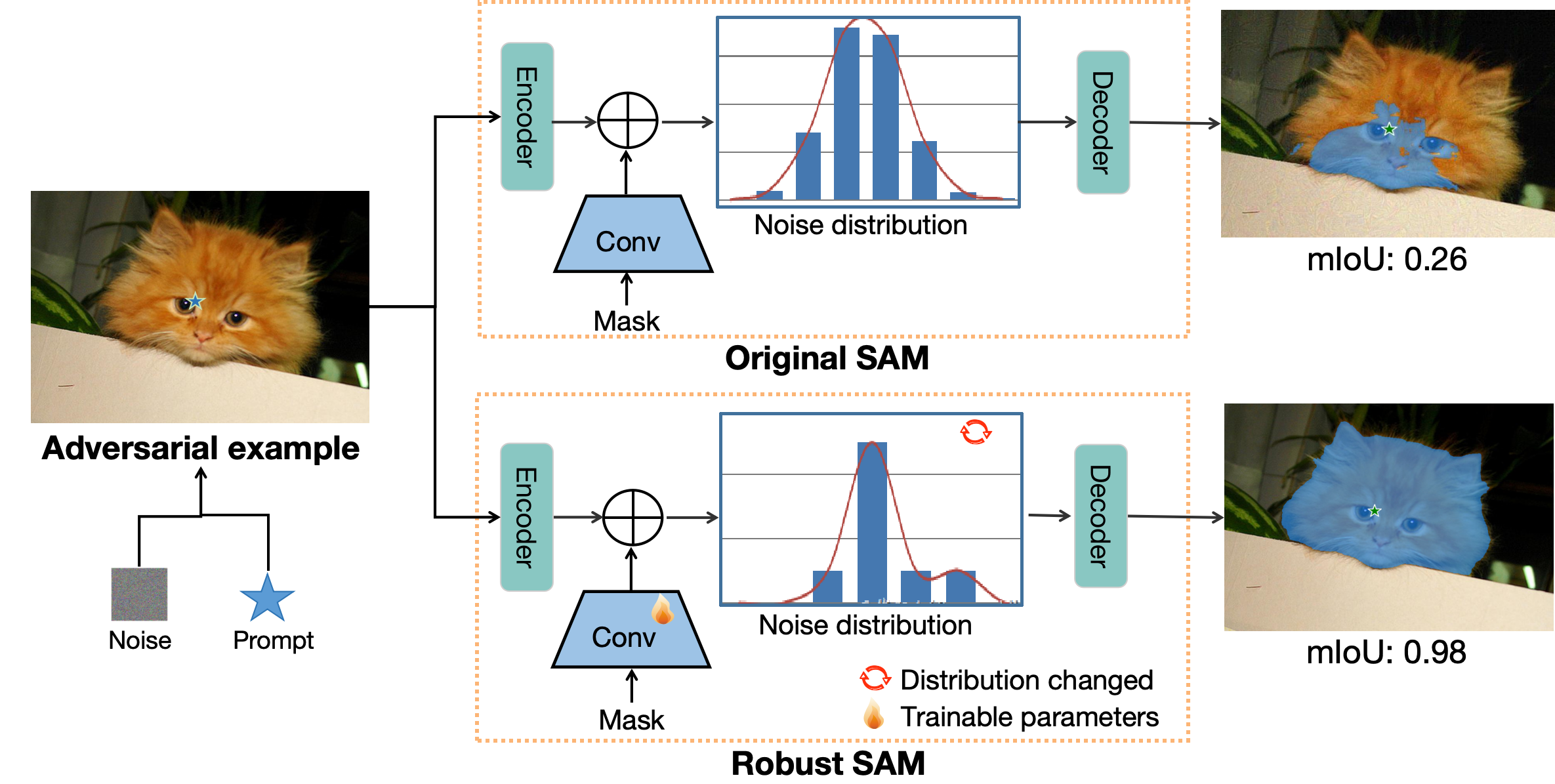}
\caption{
A comparison of the adversarial robustness of the original SAM and the proposed Robust SAM. When implementing attacks, the original SAM predicts a segmentation mask with severely compromised precision (i.e. 0.26 mIoU with the ground truth), whereas our Robust SAM maximally preserves the quality of the segmentation mask (i.e. 0.98 mIoU with the ground truth) via modified noise distribution.
}
\label{fig:different method comparison}
\end{figure}

SAM is a popular vision foundation model for image segmentation~\cite{SAM}. It predicts the object mask of an image based on box or point prompts provided by the user. SAM has been successful in this area, leading to efforts to use it for more complex tasks beyond segmentation, such as tracking and detection~\cite{trackinganything, groundedsam, Inpaintanything}. Recently, Meta pushed the boundaries further with the introduction of an even more capable and versatile successor to SAM, known as Segment Anything Model 2 (SAM 2), designed specifically for video segmentation~\cite{SAM2}.
However, existing studies have revealed that SAM is vulnerable to attacks by adversarial examples~\cite{Attack-sam, robustsam, black-sam, empirical-attack-sam}, which are subtle perturbations added to a benign image. These adversarial examples significantly degrade the segmentation performance of SAM. Since SAM has a wide range of applications, it is crucial to examine and enhance its adversarial robustness.

Recent studies have pointed out that the effectiveness of adversarial attacks against SAM is closely related to the type of input prompts~\cite{point-robust, huang2023robustness}. 
Indeed, our preliminary experiments indicate that existing attacks exhibit poor transferability across different types of prompts. For example, the PGD attack~\cite{PGD} can disrupt the segmentation mask created by a point prompt but not the one generated from a box prompt. Therefore, exploring adversarial attacks against SAM that simultaneously work on different prompt types will offer further insights into SAM's robustness.

Beyond attack methods, it is crucial to investigate the defense mechanisms of the SAM. Previous studies have proposed different approaches to defend against adversarial examples~\cite{dziugaite2016study, osadchy2017no, guo2020meets, ross2018improving}. Among these defense, full-parameter adversarial training is a commonly researched and proven effective method~\cite{ganin2016domain, bai2021recent} to increase a model's robustness. However, performing full-parameter adversarial training on large pretrained model like SAM is not only computationally expensive and prone to catastrophic forgetting~\cite{catastrophicforgetting1, catastrophicforgetting2}. Therefore, exploring few-parameter adaptation methods to bolster the robustness of large models is both meaningful and necessary.

There are some parameter adaptation schemes that aim to improve the performance of SAM in various downstream applications like medical image segmentation~\cite{samed, MedSAM} and camouflage detection~\cite{SAM-adapter-Camouflage-Shadow-more}. Yet, a direct adoption of these methods to enhance adversarial robustness faces non-trivial challenges. One major issue is that these adaptation schemes tend to prioritize precision on adversarial examples at the expense of accuracy on clean datasets.  This trade-off is especially apparent in the case of the SA1B dataset~\cite{SAM}, which serves as SAM's foundational training dataset. Therefore, it is crucial to develop a few-parameter defense method that can effectively balance SAM's robustness and accuracy.

This work focuses on enhancing the robustness of SAM against adversarial attacks while minimizing the impact on its baseline performance. This is achieved by training a limited set of parameters.
As shown in Figure~\ref{fig:different method comparison}, selectively adapting a small portion of parameters results in a more robust SAM variant (i.e., RobustSAM), which outperforms the original SAM in segmentation under attack.	
Initially, our approach begins with the design of  a cross-prompt adversarial attack to generate adversarial examples. These examples can effectively attack SAM under both point and box prompts, enabling a more comprehensive evaluation of SAM's robustness.
To further enhance the robustness of SAM, we propose a few-parameter adversarial defense on the adversarial examples.
By adapting the convolutional layers within SAM, we effectively alter the feature distribution, thereby establishing an adversarial defense against various adversarial attacks. 
To minimize the number of trainable parameters,  we employ Singular Value Decomposition (SVD)~\cite{SVDori} to reconstruct the parameters space of the convolution layers. 
By adapting only 512 singular values, we significantly improve the SAM's robustness while maximally maintaining its original segmentation ability.
\begin{itemize}
\item  We design a cross-prompt adversarial attack that updates noise to disrupt common key features of both point and box prompts.  Compared to previous SAM attacks, it can achieve higher attack transferability across different types of prompts.

\item  We develop a new SVD-based defense method for vision foundation models that enhances SAM's robustness while preserving its original performance by adapting only a small fraction of the parameters.

\item 
We validate the proposed attack and defense methods on SAM for image segmentation, demonstrating their effectiveness. Additionally, we extend the attack to SAM 2, successfully compromising its video segmentation capabilities.
\end{itemize}

\begin{table}[t]
\small
\begin{center}
\resizebox{0.47\textwidth}{!}{
\begin{tabular}{@{}cccccccc@{}}
\toprule
Methods  
 & \begin{tabular}[c]{@{}l@{}} [1]\end{tabular}   
 &\begin{tabular}[c]{@{}l@{}} [2]\end{tabular} 
 &\begin{tabular}[c]{@{}l@{}} [3]\end{tabular} 
  &\begin{tabular}[c]{@{}l@{}} [4]\end{tabular} 
&\begin{tabular}[c]{@{}l@{}} 
 [5]\end{tabular} 
 &\begin{tabular}[c]{@{}l@{}} 
 [6]\end{tabular} 
 &\begin{tabular}[c]{@{}l@{}} 
Ours\end{tabular}  \\
\midrule
\ \   New attack method &$\surd$	 &$\surd$	&$\surd$      &$\surd$     &$\surd$ 	& &$\surd$\\ 
\ \   New defense method &	 &	&      &     & 	&$\surd$ &$\surd$\\ 
\ \  Attack transferability &	 &	&      &    & 	& &$\surd$ \\ 
\ \  SAM1 \& 2  &	 &	&      &    & 	& &$\surd$ \\ 
\ \ Computation efficiency &	 &	&      &    & 	&$\surd$ &$\surd$ \\ 
\ \ Robustness tradeoff &	 &	&      &    & 	& &$\surd$ \\ \bottomrule
\end{tabular}
}
\end{center}
\caption{A comparison of representative works on SAM’s robustness. [1]-[6] represents the references~\cite{Attack-sam, robustsam, black-sam, empirical-attack-sam, redtemSAM, chen2024robustsam}, respectively.}
\label{Tab: different patch attacks} 
\end{table} 
\vspace{-0.1in}

\section{Related Work}
\paragraph{Attacking SAM.}
As the Segment Anything Model (SAM) becomes increasingly popular in the computer vision community, its robustness against adversarial attacks begins to draw greater research attention~\cite{Attack-sam, robustsam, black-sam, empirical-attack-sam, SAM_meets_surgery, redtemSAM}. Zhang et al.~\cite{Attack-sam} conduct the first investigation on how to attack SAM with adversarial examples. After this work, Qiao et al.~\cite{robustsam} conducted an extensive study to evaluate the robustness of SAM, including various corruptions, occlusions, and perturbations. Zheng et al.~\cite{black-sam} achieve targeted adversarial attacks on SAM with black-box settings. However, these methods exhibit poor attack transferability under different prompt types. In this paper, our method can impose effective attacks against SAM under either point or box prompts.

\begin{figure*}[t]
	\begin{minipage}{\linewidth}
		\centering
    \includegraphics[width=1.0\linewidth]{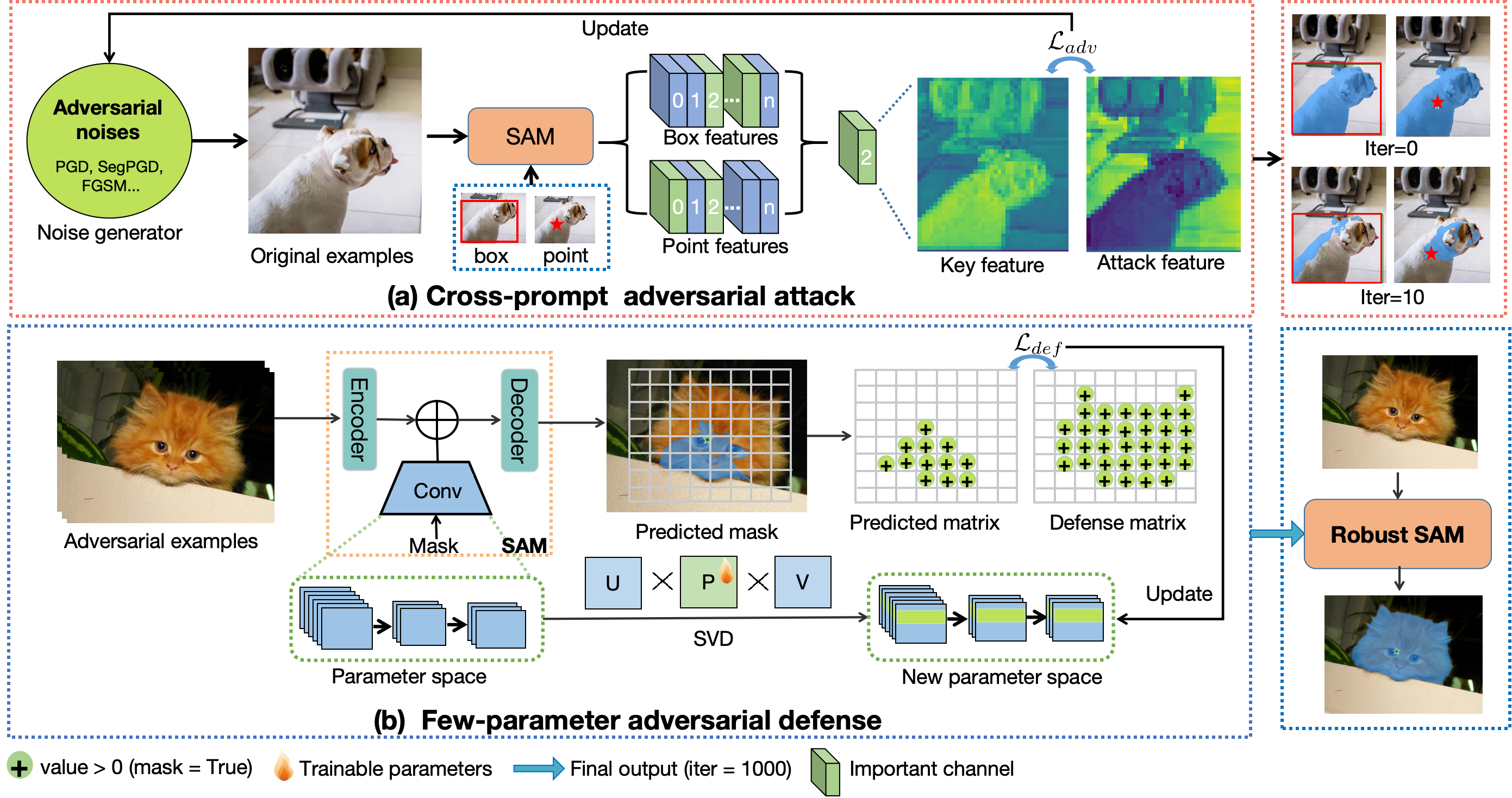}
	\end{minipage}
	\caption{The overall framework of our proposed adversarial attack and defense pipelines. (a) illustrates one iteration of the cross-prompt adversarial attack.  Original examples are perturbed by a noise generator, and these perturbed examples are then input into SAM, disrupting key features for both box and point prompts. The effectiveness and progression of the attack are demonstrated over multiple iterations. (b) depicts the adversarial defense process. Within the SAM model, the parameters of the convolutional layer (Conv) are decomposed via the singular value decomposition (SVD). 
    Only the matrix $\mathbf{P}$ in the new parameter space is updated, while other parameters in the model remain frozen.
 After the parameter adaptation, the fine-tuned SAM is validated using adversarial examples to evaluate its robustness. }
    \label{fig:optimazation framework}
\end{figure*}

\paragraph{Adapting SAM.}
Existing parameter adaptation methods for SAM~\cite{SAM-adapter-medical, SAM-adapter-Camouflage-Shadow-more, samed, SAM-COBOT, MedSAM} mainly focus on improving its performance in downstream applications, such as medical image segmentation, camouflage detection. For example, SAMed~\cite{samed} customize SAM to the medical image segmentation task using a  low-rank adaptation (LoRA)~\cite{LoRA}. Similarly, SAM-Adapter~\cite{SAM-adapter-Camouflage-Shadow-more} introduce new adapter layers into SAM's transformer blocks for camouflage detection. However, applying these adaptation methods to improve SAM's robustness can result in overfitting on adversarial datasets, which in turn may compromise its performance on clean datasets.

To address the above research gaps, we propose a comprehensive robustness framework for SAM that integrates a cross-prompt adversarial attack and an SVD-based defense method. Unlike existing approaches, our attack not only targets image and video segmentation on SAM and SAM 2 but also demonstrates higher attack transferability across different types of prompts. Additionally, our few-parameter defense method effectively preserves SAM’s original performance on clean datasets while enhancing its robustness against adversarial datasets. 
A comparison of the features and contributions of our method against prior works is provided in Table~\ref{Tab: different patch attacks}.

\section{Methodology}

\subsection{Cross-Prompt Adversarial Attack}\label{cross-prompt attack}

\subsubsection{Attack Define.}
Previous attacks on image segmentation models have primarily aimed to alter the category labels of predicted segmentation masks~\cite{attacksegmentation, patchattacksegmentation}. 
However, SAM’s predicted masks do not include categorical labels; instead, they indicate only the shape of plausible objects. As such, our attack is formulated with the objective of disrupting the shape of the segmentation masks generated by SAM.  
Moreover, as SAM works under the guidance of two types of user-provided prompt (i.e., point and box prompts), the proposed attack shall generalize to both types of prompts.

Consider a fixed set of images \( \mathcal{X} = \{\mathbf{X}^1, \mathbf{X}^2, \ldots, \mathbf{X}^N\} \), where \( N \in \mathbb{N} \) is the total number of images. For any two masks \( A \) and \( B \) corresponding to the same image \( \mathbf{X}^{i} \), their mean Intersection over Union (mIoU) is defined as:
\begin{equation} \label{mIoU}
    \operatorname{mIoU}(A, B) = \frac{|A \cap B|}{|A \cup B|}
\end{equation}
The indicator function \(I(i)\) for image \(X^{i}\) is defined as:
\begin{equation} 
    I(i) = \begin{cases} 
        1 & q^{point}(i) < 0.5
           \; \& \;
         q^{box}(i) < 0.5 \\ 
        0 & \text{otherwise} 
    \end{cases}
\end{equation}
where 
\begin{eqnarray}
    q^{point}(i)= \operatorname{mIoU}(\operatorname{S}^{point}(\mathbf{X}_{adv}^{i}), \operatorname{S}^{point}(\mathbf{X}^{i})), \\
    q^{box}(i)=\operatorname{mIoU}(\operatorname{S}^{box}(\mathbf{X}_{adv}^{i}), \operatorname{S}^{box}(\mathbf{X}^{i})). 
\end{eqnarray}
Here, $\mathbf{X}_{adv}^{i}$ is the adversarial example for benign image $\mathbf{X}^{i}$;
\( \operatorname{S}^{point}(\mathbf{X}_{adv}^{i}) \) and \( \operatorname{S}^{box}(\mathbf{X}_{adv}^{i}) \) denote SAM's predicted masks for \(\mathbf{X}_{adv}^{i}\) under point and box prompts, respectively; likewise, \( \operatorname{S}^{point}(\mathbf{X}^{i}) \) and \( \operatorname{S}^{box}(\mathbf{X}^{i}) \) denote SAM's predicted masks for \(\mathbf{X}^{i}\) under point and box prompts, respectively. 
Essentially, the indicator function defines an attack to be successful only if the mIoU for the predicted masks of the original image and its corresponding adversarial examples concurrently fall below 0.5 under the point and box prompts.
Based on the indicator function, we define the attack success rate (ASR) as $ASR =\frac{1}{N}\sum_{i=1}^{N} I(i)$.

\subsubsection{Attack Motivation.}
Existing works~\cite{wang2023improving, wang2021feature} indicate that disrupting the important features can dramatically reduce the model accuracy. Building on this insight, \textbf{our cross-prompt attack is designed to affect the common key features of the model across both point and box prompts, thereby enhancing the attack transferability.} To achieve this, we propose a novel key feature selection algorithm, detailed as follows.

\begin{figure}[t]
\centering
\includegraphics[scale=0.28]{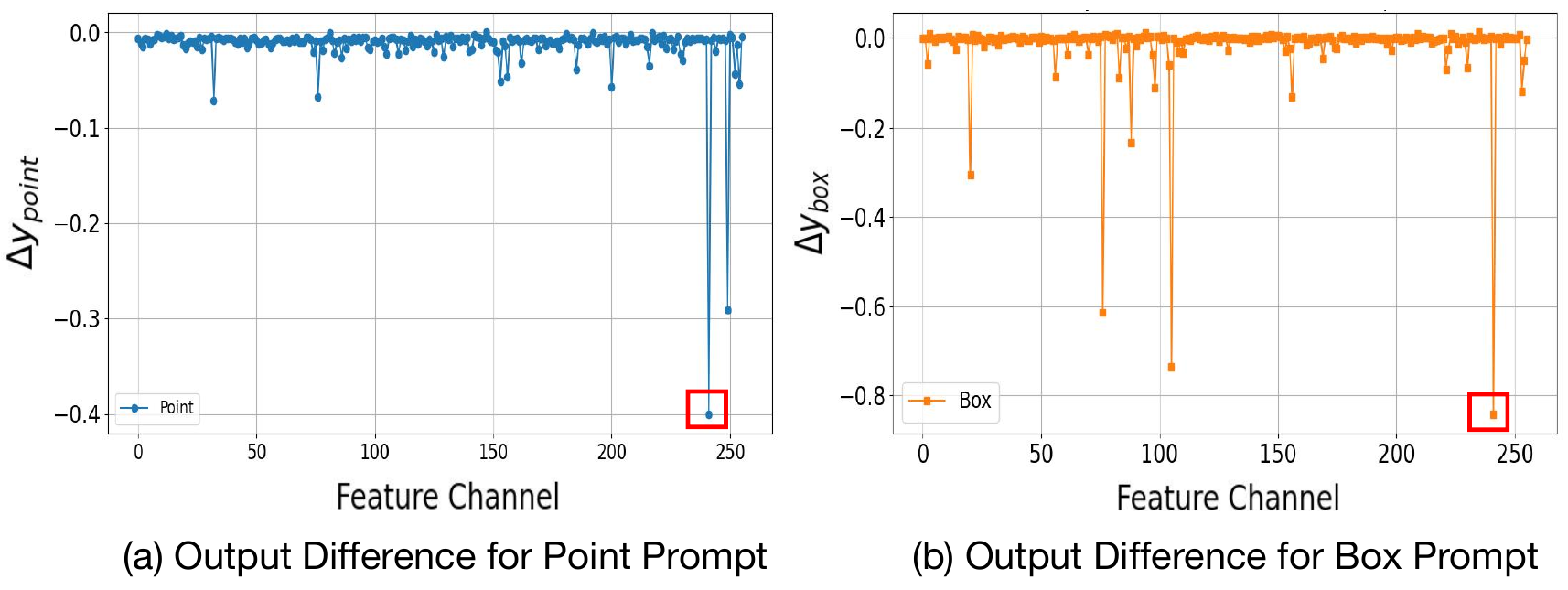}
\caption{Demonstrations of output differences for point and box prompts. It illustrates the negative impact on SAM's output when the feature map of each individual channel is set to zero (i.e., $f_i(\mathbf{X}) = 0$). \textcolor{red}{Red boxes} highlight the key features that have the greatest impact on mIoU for the point and box prompts.} 
\label{Demonstration of output diffenreces}
\end{figure}

\subsubsection{Key Feature Selection.} SAM employs a Vision Transformer as its encoder to extract global image features and utilizes a decoder to convert these features into precise segmentation masks.
Given an image $\mathbf{X}$, the SAM encoder outputs a set of features $\mathcal{F} = \{f_i(\mathbf{X})\}$, where $i \in \{1, 2, \dots, N\}$, with N representing the total number of channels. To evaluate the importance of each feature, we generate modified feature sets $\{\tilde{\mathcal{F}}_i\}$, where each set $\tilde{\mathcal{F}}_i$ is constructed by setting feature $f_i(\mathbf{X})$ to 0 while keeping all other features unchanged. For each channel $i$, we calculate the output differences for the point prompt and box prompt:
\begin{equation}
    \Delta y_{\text{point}, i} = \text{Decoder}_{\text{point}}(\mathcal{F}) - \text{Decoder}_{\text{point}}(\tilde{\mathcal{F}}_i),
\end{equation}
\begin{equation}
    \Delta y_{\text{box}, i} = \text{Decoder}_{\text{box}}(\mathcal{F}) - \text{Decoder}_{\text{box}}(\tilde{\mathcal{F}}_i),
\end{equation}

The channels of common key features are calculated by:
\begin{equation}
    C_{\text{point}} = \text{TopK}(\Delta y_{\text{point}, i}), \quad C_{\text{box}} = \text{TopK}(\Delta y_{\text{box}, i}),
\end{equation}
\begin{equation}
    C_{\text{common}} = C_{\text{point}} \cap C_{\text{box}}, 
\end{equation}

where the function $\text{TopK}$ selects top $K$ feature channels with the largest output differences. Figure~\ref{Demonstration of output diffenreces} provides a specific example, illustrating the output differences for both point and box prompts when each feature channel is set to zero.

\subsubsection{Attack Pipeline.}

As shown in Figure 2(a), SAM extracts two types of features—point features and box features—from an input image. Although these two types of features are fundamentally similar, they contribute differently to SAM’s output depending on the prompt type. Next, we use our key feature selection algorithm to extract the common key features from both types. Subsequently, we define $\mathbf{A}$ as an attack feature that uses negative values to disrupt the key features, leading to less effective feature representations. Finally, we optimize the adversarial noise using \(\mathcal{L}_{adv}\), which quantifies the discrepancy between the key features and the attack features. $\mathcal{L}_{adv}$ is given by:
\begin{eqnarray}
    \mathcal{L}_{adv} = \text{MSE}(\mathbf{A}_{i}, f_i(\mathbf{X} + \delta)), \quad i \in C_{\text{common}}
\end{eqnarray}

where \(f_i(\mathbf{X} + \delta)\) represents the feature map of the \(i\)-th channel produced by SAM’s encoder given the adversarial image \(\mathbf{X} + \delta\). The adversarial noise \(\delta\) is updated in the next iteration as 
$\delta^* = \text{Clip}_{\mathbf{X}, \epsilon}(\delta + \alpha \cdot \text{sign}(\nabla_\delta \mathcal{L}_{adv}))$
with a step size \(\alpha\). The function \(\text{Clip}_{\mathbf{X}, \epsilon}(\cdot)\) constrains the noise within \([- \epsilon, \epsilon]\) and ensures the adversarial image \(\mathbf{X} + \delta\) remains within the valid range of [0, 255].

\subsection{Few-Parameter Adversarial Defense}
\paragraph{Defense Define.}
The objective of our defense method is to enhance SAM's robustness, enabling it to recover the correct masks for adversarial examples. 
We define a set of adversarial examples \( \mathcal{X} = \{\mathbf{X}_{adv}^1, \mathbf{X}_{adv}^2, \ldots, \mathbf{X}_{adv}^N\} \), where \( N \in \mathbb{N} \) is the total number of adversarial examples in the set. Our goal is to train a more robust version of SAM (denoted by RobustSAM) that maximizes
$ \operatorname{mIoU}(\operatorname{RS}^{p}(\mathcal{X}), \mathcal{GT}) $ under different types of prompts $p$, where $p \in [point, box]$. $\operatorname{RS}^{p}(*)$ is the output masks of the RobustSAM under prompt $p$, and $\mathcal{GT}$ denotes the set of ground truth masks corresponding to the adversarial examples in $\mathcal{X}$.

\begin{figure}[t]
\centering
\includegraphics[scale=0.3]{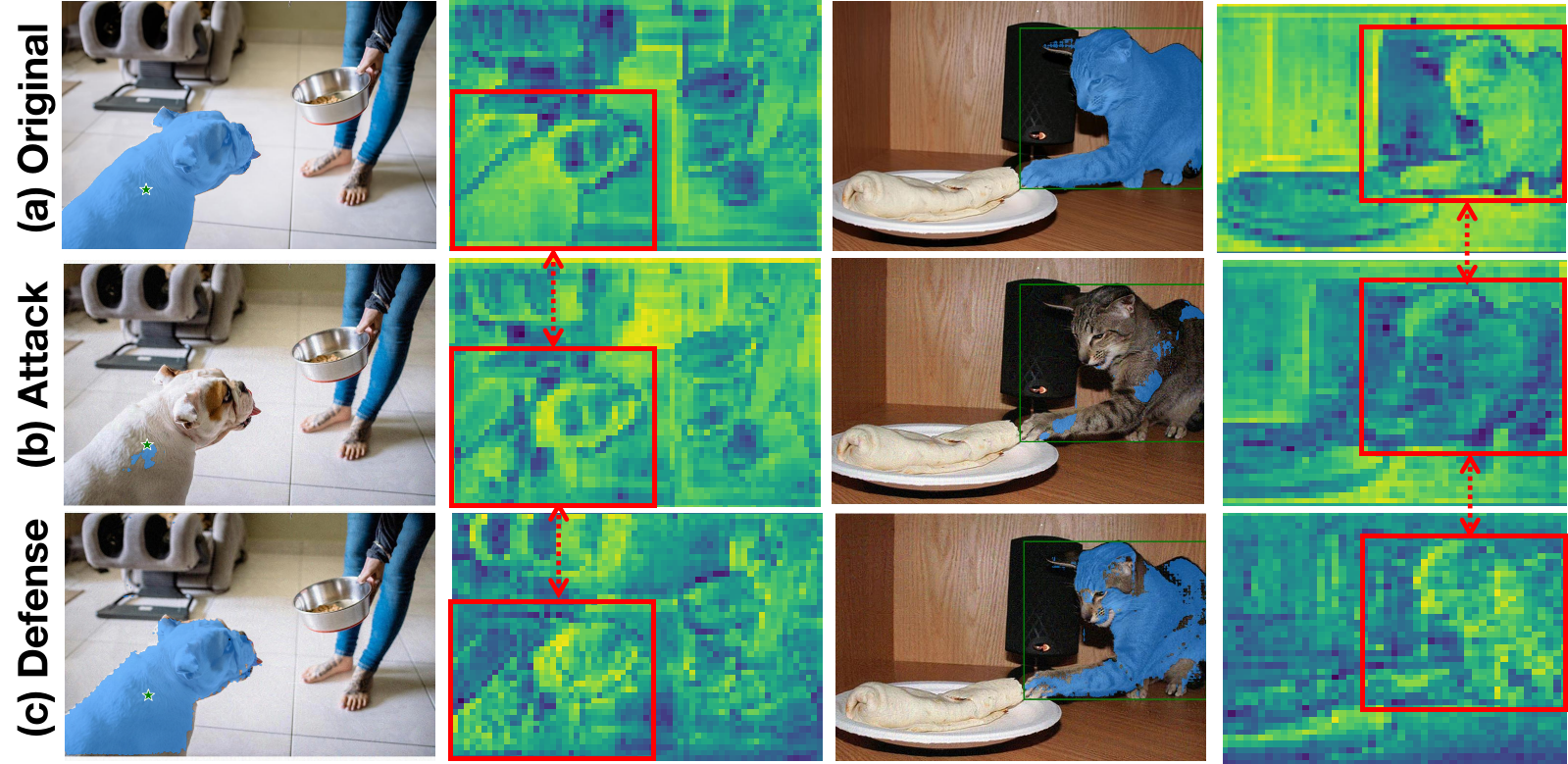}
\caption{Effects of attack and defense on the intermediate features in SAM. Top to bottom: feature maps of (a) a clean image, (b) an image corrupted by adversarial noise, and (c) an image corrupted by adversarial noise but defended by our proposed defense method. } 
\label{fig:aggregation}
\end{figure}

\paragraph{Defense Motivation.}
Existing works~\cite{xie2019feature, ilyas2019adversarial, long2024papmot} indicate that adversarial noise, while almost imperceptible when added into the pixel space, induces apparent perturbations in the feature space. As illustrated in Figure~\ref{fig:aggregation}, the feature map of a clean image (a) differs significantly from that of an image corrupted by adversarial noise (b), where the texture features of the ``dog" and ``cat" are obscured by the added noise.
As such, \textbf{the motivation behind our defense method is to mitigate the noise impact by altering the feature distribution of the adversarial example, with minimal parameter adaptation.} Figure~\ref{fig:aggregation} (c) demonstrates that the proposed RobustSAM maintains the segmentation quality of adversarial examples by adjusting the feature distribution.

\paragraph{Trainable Parameters Selection.} 
To achieve our goal of improving SAM's robustness with minimal parameter adaptations, selecting the appropriate trainable parameters is crucial.
The convolutional layers, as the final stage of SAM's feature extraction network (i.e., encoder), play a important role to adjusting feature distribution. These layers not only reduce the dimensionality of the feature space but also contain relatively few parameters ($\approx$ 600K), accounting for just 1/100th of SAM's total parameters.

Additionally, we can further compress these trainable parameters using Singular Value Decomposition (SVD). In our approach, the parameters of the convolutional layers are decomposed into three matrices: $\mathbf{U}$, $\mathbf{P}$, and $\mathbf{V}$. $\mathbf{U}$ and $\mathbf{V}$ are orthogonal matrices that represent the bases for the input and transformed spaces, respectively, with each column capturing specific patterns or features in the input data. $\mathbf{P}$ is a diagonal matrix containing singular values that scale these bases.
The decomposition process can be formulated as:
\begin{eqnarray}
\mathbf{W} =\sum_{i=1}^{d}  \mathbf{u}_{i}p_{i}\left(\mathbf{v}_{i}\right)^{T} 
 =\mathbf{U P V}^{T}, 
\end{eqnarray}

where $\mathbf{W} \in \mathbb{R}^{d \times k}$ denotes the parameter space of a convolutional layer with $d$ input channels and $k$ output channels. 
$\mathbf{U} \in \mathbb{R}^{d \times d}$  
and 
$\mathbf{V} \in \mathbb{R}^{k \times d}$  are the basis matrices while $\mathbf{P} = \mathbf{diag}(p_{1}, p_{2}, \ldots, p_{i}, \ldots, p_{d})$ is the diagonal matrix adjusting these bases.
To sum up, adjusting the matrix 
$\mathbf{P}$ is equivalent to adjusting the entire parameter space of the convolutional layers in specific directions. 
Our defense improves the adversarial robustness of SAM by training only the diagonal matrix $\mathbf{P}$.

\paragraph{Defense Pipeline.}
As illustrated in Figure~\ref{fig:optimazation framework}(b), RobustSAM encompasses two key processes: parameter adaptation and parameter decomposition.
During parameter adaptation process, SAM takes as input an adversarial example and outputs a prediction mask for it. 
To encourage correct mask prediction by SAM, we create a defense matrix with positive values that correspond to the pixel locations of the ground truth masks. Based on the mean squared error (MSE), we then calculate a loss $\mathcal{L}_{def}$ between the predicted matrix and the defense matrix. 
During the parameter decomposition process, we focus on the convolutional layers in SAM's feature extraction network.
Rather than adapting all parameters, we apply singular value decomposition (SVD) to decompose the convolutional layer's parameter space into three matrices $\mathbf{U}$, $\mathbf{P}$, and $\mathbf{V}$. Among them, we specifically set diagonal matrix $\mathbf{P}$ as the trainable parameters while other parameters in the model remain frozen. 
The optimization objective for the proposed adversarial defense is given by:
\begin{gather}
            \mathcal{L}_{def} = \operatorname{MSE}\left(\mathbf{D},  
    \operatorname{RS}^{p}\left(\mathbf{X}_{adv}\right)\right),    \\
    \mathbf{P}^{*}=\arg \min _{\mathbf{P}} \mathcal{L}_{def},
\end{gather}
where $\operatorname{RS}^{p}(*)$ represents the output masks of RobustSAM under different types of prompts $p$, 
$\mathbf{D}$ is the defense matrix with positive values according to the ground truth masks, and  $\mathbf{P}$ is a diagonal matrix in the new parameters space of RobustSAM.

Our defense improves the adversarial robustness of SAM by training only diagonal matrix $\mathbf{P}$.
We stress that this design enjoys the following three merits: (1) \textit{Varied feature distribution}. The matrix $\mathbf{P}$ acts as a coefficient matrix that directly adjusts the feature space, allowing us to modify the feature distribution and reduce the impact of adversarial noise. (2) \textit{Balanced robustness and performance}. By training only $\mathbf{P}$ and keeping the foundational matrices $\mathbf{U}$ and $\mathbf{V}$) unchanged, we maintain the core feature space while adapting the model's response to adversarial examples, preventing overfitting to noise.
(3) \textit{Reduced training cost}. Training only $\mathbf{P}$ significantly lowers the training cost, as we improve SAM's robustness with only 512 parameters, compared to the 93735K parameters of the model or 600K of the convolutional layers.

\begin{figure*}[t]
	\begin{minipage}{\linewidth}
	\centering
        \includegraphics[width=1.0\linewidth]{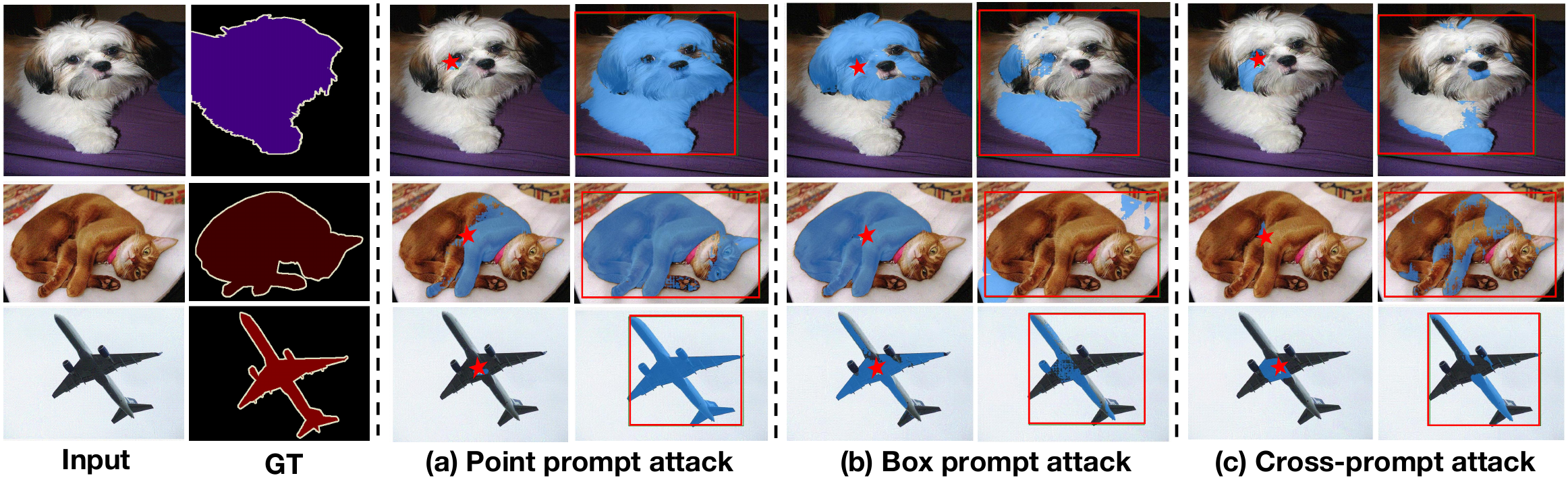}
	\end{minipage}
	\caption{Illustrations of various adversarial attacks against SAM by employing both point and box prompts. An attack is considered successful if it removes more than 50\% of the ground truth masks for either point or box prompts. (a) illustrates that the point prompt attacks effectively compromise the ground truth masks under point prompts but are ineffective when encountering box prompts. (b) depicts the box prompt attacks that successfully disrupt the ground truth masks with point prompts but fail to attack under point prompts. 
    (c) showcases that our cross-prompt attack successfully degrades the performance for both point and box prompts simultaneously. }
	\label{fig:comparison of attacks}
\end{figure*}

\section{Experiments}

\subsection{Experimental Setup}
\paragraph{Datasets and Metrics.} 
To evaluate the robustness of SAM under different types of prompts (i.e., point and box prompts), we randomly sample 2000 images from the SA1B~\cite{SAM}, VOC~\cite{VOCdataset}, COCO~\cite{COCOdataset}, and DAVIS~\cite{davis} datasets. 
Each image is paired with a ground truth segmentation mask of a single object, along with corresponding prompts.
The VOC dataset is split into 70\% for training and 30\ for evaluation. SA1B and VOC are used to assess SAM's robustness under independent and identically distributed (i.i.d.) conditions, while COCO, DAVIS is used for out-of-distribution (o.o.d.) conditions.
For attack, we use the attack success rate (ASR) and the mean Intersection over Union (mIoU) as the evaluation metric. 
For defense, mIoU is used as the evaluation metric.

\paragraph{Baselines.}
We choose two victim models for our experiments: SAM~\cite{SAM} and SAM 2~\cite{SAM2}.
Existing attacks against SAM can be categorized based on the type of prompt inputs: point prompt attack (PPA)~\cite{Attack-sam,SAM_meets_surgery} and box prompt attack (BPA)~\cite{robustsam}.  These methods are variants of gradient-based adversarial attacks, such as PGD~\cite{PGD} and FGSM~\cite{FGSM}. For a fair comparison, PPA, BPA and our cross-prompt attack are configured with the same attack strength and number of iterations. For defense comparison, we introduce various parameter adaptation schemes to improve the robustness, including SAM-LoRA~\cite{samed}, SAM-Adapter~\cite{SAM-adapter-medical}, MedSAM~\cite{MedSAM}, SAM-COBOT~\cite{SAM-COBOT} and FacT~\cite{jie2023fact}. The main difference among these approaches is the selection of parameters within SAM for adaptation.
For more details on these attack and defense methods, please refer to the Appendix.

\paragraph{Implementation Details.}
For the attack setting, we set the total number of iteration steps to 20, and perturbation intensity $\epsilon$ to 16/255 for PPA, BPA, and our cross-prompt attack. The attack feature $\mathbf{A}$ are set to the negative values of the key features, and $K$ in TOPK function is set to 5. 
For all few-parameter adaptation methods, we randomly sample 70\% of the adversarial examples in the VOC dataset for adapting SAM. 
Our training employs the Adam optimizer~\cite{adam}. The initial learning rate is set to \(1.0 \times 10^{-3}\), and the weight decay is \(5 \times 10^{-5}\) with one image per mini-batch. The number of training epochs is set to 500.

\begin{table}[t]
\renewcommand{\arraystretch}{1.3} 
\resizebox{0.47\textwidth}{!}{
\begin{tabular}{lclccclccclccl}
\hline
 &  &  & \multicolumn{3}{c}{COCO} &  & \multicolumn{3}{c}{VOC} &  & \multicolumn{3}{c}{SA1B} \\ \cline{4-6} \cline{8-10} \cline{12-14} 
 & $\epsilon$ &  & 4 & 8 & 16 &  & 4 & 8 & 16 &  & 4 & 8 & 16 \\ \hline
\multicolumn{1}{l|}{} & PPA &  & 1.4 & 2.8 & 11.4 &  & 2.6 & 5.4 & 8.3 &  & 0.2 & 2.5 & 3.3 \\
\multicolumn{1}{l|}{} & BPA &  & 1.9 & 12.8 & 35.7 &  & 2.4 & 19.1 & 44.4 &  & 14.2 & 33.4 & 44.5 \\
\multicolumn{1}{l|}{\multirow{-3}{*}{\begin{tabular}[c]{@{}l@{}}\rotatebox{90}{SAM}\\ \end{tabular}}} & \cellcolor[HTML]{EFEFEF}CPA (Ours) &  \cellcolor[HTML]{EFEFEF}& \cellcolor[HTML]{EFEFEF}\textbf{5.7} & \cellcolor[HTML]{EFEFEF}\textbf{15.7} & \cellcolor[HTML]{EFEFEF}\textbf{60.2} &  \cellcolor[HTML]{EFEFEF}& \cellcolor[HTML]{EFEFEF}\textbf{6.2} & \cellcolor[HTML]{EFEFEF}\textbf{30.3} & \cellcolor[HTML]{EFEFEF}\textbf{62.9} &  \cellcolor[HTML]{EFEFEF}& \cellcolor[HTML]{EFEFEF}\textbf{20.4} & \cellcolor[HTML]{EFEFEF}\textbf{36.2} & \cellcolor[HTML]{EFEFEF}\textbf{75.6} \\ \hline
\multicolumn{1}{l|}{} & PPA &  & 0 & 0.3 & 5.9 &  & 0.2 & 0.5 & 3.1 &  & 0 & 0.2 & 0.3 \\
\multicolumn{1}{l|}{} & BPA &  & 0 & 6.6 & 18.5 &  & 0 & 9.8 & 20.0 &  & 7.5 & 13.2 & 21.5 \\
\multicolumn{1}{l|}{\multirow{-3}{*}{\begin{tabular}[c]{@{}l@{}}\rotatebox{90}{SAM 2}\\ \end{tabular}}} & \cellcolor[HTML]{EFEFEF}CPA (Ours) &  \cellcolor[HTML]{EFEFEF}& \cellcolor[HTML]{EFEFEF}\textbf{1.9} & \cellcolor[HTML]{EFEFEF}\textbf{8.5} & \cellcolor[HTML]{EFEFEF}\textbf{32.8} &  \cellcolor[HTML]{EFEFEF}& \cellcolor[HTML]{EFEFEF}\textbf{2.2} & \cellcolor[HTML]{EFEFEF}\textbf{18.7} & \cellcolor[HTML]{EFEFEF}\textbf{34.4} &  \cellcolor[HTML]{EFEFEF}& \cellcolor[HTML]{EFEFEF}\textbf{12.0} & \cellcolor[HTML]{EFEFEF}\textbf{19.1} & \cellcolor[HTML]{EFEFEF}\textbf{36.9} \\ \hline
\end{tabular}
}
\caption{Comparisons of ASR with point prompt attack (PPA), box prompt attack (BPA), and our cross-prompt attack (CPA) on multiple datasets. $\epsilon$ denotes the perturbation intensity.}
\label{ASR of cross}
\end{table}

\begin{table*}[t]
\renewcommand{\arraystretch}{1.3}
\centering
\resizebox{1.0\textwidth}{!}{
\begin{tabular}{clccclccclccclccc}
\hline
 &  & \multicolumn{3}{c}{VOC} &  & \multicolumn{3}{c}{COCO} &  & \multicolumn{3}{c}{SA1B} &  & \multicolumn{3}{c}{DAVIS} \\ \cline{3-5} \cline{7-9} \cline{11-17} 
 &  & BPA & PPA & CPA(ours) &  & BPA & PPA & CPA(ours) &  & BPA & PPA & CPA(ours) &  & BPA & PPA & CPA(ours) \\ \hline
Baseline$^\dag$ &  & 16.9 & 29.5 & 6.4 &  & 11.7 & 26.9 & 5.9 &  & 17.2 & 23.7 & 8.1 &  & 18.3 & 30.1 & 7.5 \\ \hline
MedSAM~\cite{MedSAM} &  & 22.1 & 24.3 & 8.7 &  & 13.2 & 27.1 & 9.4 &  & 19.5 & 24.0 & 11.8 &  & 23.5 & 26.7 & 10.2 \\
SAM-Adapter~\cite{SAM-adapter-Camouflage-Shadow-more} &  & 27.1 & 34.4 & 25.3 &  & 22.3 & 38.0 & 18.1 &  & 27.5 & 38.1 & 16.9 &  & 29.4 & 36.5 & 28.7 \\
SAM-LoRA~\cite{samed} &  & 32.8 & 44.0 & 27.0 &  & 33.9 & 45.1 & 28.1 &  & \textbf{44.3} & \textbf{57.5} & 31.9 &  & 33.6 & 45.3 & 29.3 \\
SAM-COBOT~\cite{SAM-COBOT} &  & 31.1 & 46.3 & 30.4 &  & 30.2 & 41.2 & 23.8 &  & 40.9 & 58.7 & 34.8 &  & 32.1 & 48.9 & 33.7 \\
\rowcolor[HTML]{EFEFEF}
RobustSAM (Ours) &  & \textbf{39.5} & \textbf{50.8} & \textbf{35.0} &  & \textbf{37.2} & \textbf{47.1} & \textbf{29.6} &  & 41.5 & 52.2 & \textbf{36.0} &  & \textbf{40.2} & \textbf{52.7} & \textbf{37.9} \\ \hline
\end{tabular}
}
\caption{Comparison of mIoU with different parameter adaptation methods on adversarial datasets. ``Baseline$^\dag$" denotes the original performance of the undefended SAM. Visualizations are provided in the Appendix.}
\label{Table:Comparison with defense methods}
\end{table*}

\subsection{Evaluation of Cross-Prompt Adversarial Attack}
Figure~\ref{fig:comparison of attacks} qualitatively evaluates the performance of the proposed cross-prompt attack method.  
PPA (a) can effectively disrupt the mask predicted by SAM under point prompts (i.e., the ground truth masks of ``dog", ``cat" are removed under point prompts). However, PPA fails to achieve the same level of disruption under box prompts. 
BPA (b) effectively removes more than 50\% of the ground truth mask, but fails to do so under a point prompt.
In contrast, our cross-prompt attack (c) successfully disrupts the ground truth masks under both point and box prompts, outperforming PPA and BPA. This indicates a significant improvement in adversarial attacks against SAM.

\begin{figure}[t]
\centering
\includegraphics[scale=0.42]{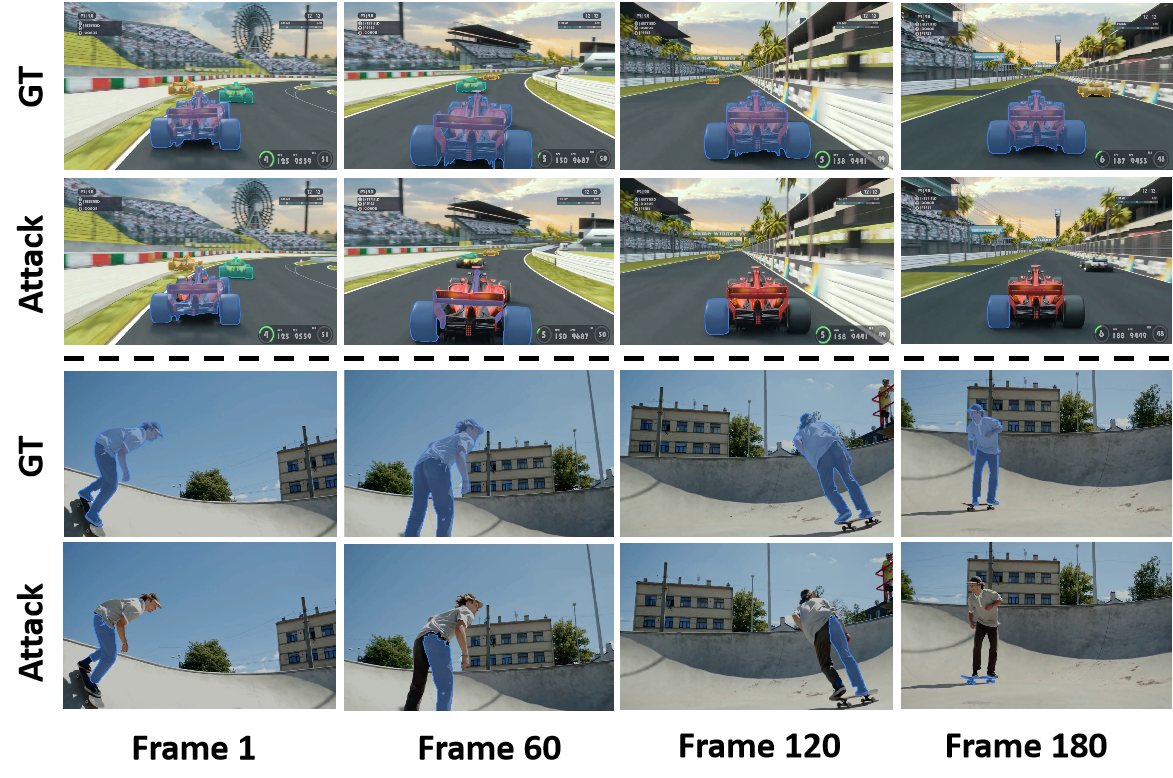}
\caption{Illustration of our cross-prompt attack against the video segmentation on SAM 2 (refer to Supplementary Materials for a video demo).}
\label{attacksam2}
\end{figure}

Figure~\ref{attacksam2} illustrates our cross-prompt attack targeting video segmentation on SAM 2~\cite{SAM2}. It demonstrates that our attack can disrupt the temporal consistency of video segmentation. For example, it misleads SAM 2 into incorrectly tracking the race car's mask as its wheels and causes the man's mask to be tracked as the skateboard. Therefore, our attack effectively compromises the expected continuity and accuracy of video segmentation.

Table~\ref{ASR of cross} quantitatively evaluate the performance of the proposed cross-prompt attack method on SAM~\cite{SAM} and SAM 2~\cite{SAM2}.
Compared to point prompt attack (PPA) and box prompt attack (BPA), the cross-prompt attack exhibits a higher attack success rate across all datasets and attack intensities (i.e., $\epsilon=4/255, 8/255, 16/255$).
On SAM, the cross-prompt attack achieves the highest success rates of 60.2\%, 62.9\%, and 75.6\% on the COCO, VOC, and SA1B datasets, respectively, significantly outperforming other methods. On SAM 2, although the overall success rates are lower, the cross-prompt attack still shows higher success rates than other attacks, especially at higher attack intensities (i.e., $\epsilon=16/255$), with success rates of 32.8\%, 34.4\%, and 36.9\%. 
This indicates that our cross-prompt attack has strong generalizability across different SAM models and datasets.

\subsection{Evaluation of Few-Parameter Adversarial Defense}

\begin{table}[t]
\centering
\resizebox{0.47\textwidth}{!}{
\begin{tabular}{cclclclclclc}
\toprule
 & \multicolumn{3}{c}{VOC} &  & \multicolumn{3}{c}{COCO} &  & \multicolumn{3}{c}{SA1B} \\ \cmidrule{2-4} \cmidrule{6-8} \cmidrule{10-12} 
 & Point &  & Box &  & Point &  & Box &  & Point &  & Box \\ \midrule
Baseline$^\dag$ & 53.5 &  & 87.1 &  & 38.1 &  & 78.8 &  & 32.0 &  & 94.0 \\ \midrule
MedSAM & 23.4 &  & 38.2 &  & 6.1 &  & 11.5 &  & 3.9 &  & 8.6 \\
SAM-Adapter & 62.7 &  & 57.9 &  & 39.5 &  & 32.6 &  & 13.1 &  & 49.4 \\
     
SAM-LoRA & 66.5 &  & 63.4 &  & 50.6 &  & 49.4 &  & 15.6 &  & 58.0 \\
SAM-COMBOT & 68.6 &  & 61.1 &  & 53.5 &  & 42.4 &  & 19.7 &  & 59.4 \\
\rowcolor[HTML]{EFEFEF}
RobustSAM (Ours) & \textbf{71.3} &  & \textbf{80.9} &  & \textbf{54.8} &  & \textbf{64.2} &  & \textbf{28.7} &  & \textbf{74.0} \\ \bottomrule
\end{tabular}
}
\caption{Comparison of mIoU with different parameter adaptation methods on clean datasets. ``Baseline$^\dag$" denotes the original performance of SAM. ``Point" and ``Box" represent the model outputs under point prompts and box prompts, respectively.}
\label{comparison of clean datasets}
\end{table}

Table~\ref{Table:Comparison with defense methods} compares the effectiveness of various parameter adaptation methods against adversarial attacks, including BPA, PPA, and our cross-prompt attack (CPA). As shown, RobustSAM achieves the highest mIoU, manifesting its superiority over existing methods (i.e., MedSAM, SAM-Adapeter, FacT, SAM-LoRA, SAM-COBOT) across VOC, COCO, SA1B and DAVIS datasets. We observe that under BPA and PPA attacks on the SA1B dataset, the mIoU of RobustSAM is slightly lower than that of SAM-LoRA. This may be attributed to LoRA's low-rank adaptations~\cite{LoRA}, which might be more effective at capturing the feature characteristics specific to the SA1B dataset. Nonetheless, RobustSAM surpasses SAM-LoRA by a significant margin on the VOC, COCO, DAVIS datasets. Additionally, we found that MedSAM underperforms compared to other methods, likely due to its adaptation of both the encoder and decoder parameters in SAM, leading to severe overfitting. In conclusion, experiments demonstrates that enhancing the robustness of SAM does not depend on the number of trainable parameters, but rather on which parameters are adapted. This finding aligns with our previous defense theory, where we can enhanced robustness by modifying the feature distribution using a small fraction of the parameters.

Table~\ref{comparison of clean datasets} presents the fundamental performance comparison on the clean datasets after applying various parameter adaptation methods. The results demonstrate that RobustSAM can maximally preserve its fundamental segmentation ability on clean datasets, whereas existing methods tend to accomplish the robustness objective at the cost of accuracy of clean datasets. For instance, on the VOC dataset under Box prompts, SAM-LoRA reduces the baseline mIoU from 87.1 to 63.4 on the clean dataset, whereas our method results in a smaller decrease, from 87.1 to 80.9. This indicates that our defense successfully improves robustness on adversarial datasets while preserving strong segmentation performance on clean datasets.

Table~\ref{tab:complexity} presents the computational comparison of different parameter adaptation method.  The results show that the complexity of our method is significantly lower than that of other parameter adaptation methods. For example, the parameters of RobustSAM are 600 times less than that of SAM-Adapter. This significant reduction in overhead stems from that our RobustSAM emphasizes adapting the core parameters, rather than adjusting a large number of parameters.

\begin{table}[t]
\tabcolsep=0.075cm
\small
\centering
\resizebox{0.4\textwidth}{!}{
\begin{tabular}{c|cc}
\toprule 
Method& Params (K) & Time (Fps) \\ \midrule
{ Baseline$^\dag$}&{93735}&3.3  \\ \midrule
MedSAM~\cite{MedSAM}&93729 &3.3 \\ 
SAM-Adapter~\cite{SAM-adapter-medical}&351.5&6.2\\
SAM-LoRA~\cite{samed}&144&5.2 \\ 
SAM-COBOT~\cite{SAM-COBOT} &1.3&10.8 \\
RobustSAM (Ours)&\textbf{0.5}&\textbf{11.5}  \\
\bottomrule 
\end{tabular}
}
\caption{Computation efficiency analysis for different parameter adaptation methods. ``Baseline$^\dag$'': adapting SAM's all parameters.}
\label{tab:complexity}
\end{table}

\section{Conclusion}
In this paper, we propose a comprehensive adversarial robustness framework for SAM including both adversarial attack and defense methods. 
The proposed cross-prompt attack effectively reveals the vulnerabilities of SAM and SAM 2 across different prompt types.
To counter these vulnerabilities, we propose RobustSAM, a few-parameter defense approach that involves training a small subset of parameters in SAM's feature extraction network. 
By applying  SVD, we further control the number of trainable parameters, ensuring a balance between enhanced robustness and maintenance of the baseline performance of SAM.
Our experimental results demonstrate that RobustSAM significantly boosts SAM's robustness against adversarial attacks by adapting only 512 parameters. 
We expect that this work bring new insights into existing research on the robustness of SAM and serve as a baseline that facilitates further research in this direction.

\noindent \textbf{Acknowledgements.} This work was supported in part by NSFC (62322113, 62376156, 62401361).

\bibliography{main}

\end{document}